\documentclass{uncecomp2019}

\usepackage{amsmath}
\usepackage{amssymb}
\usepackage{subcaption}

\title{MODEL INFERENCE FOR ORDINARY DIFFERENTIAL EQUATIONS BY PARAMETRIC POLYNOMIAL KERNEL REGRESSION}

\author{David K. E. Green$^{1,2}$, Filip Rindler$^{1,2}$}

\heading{David K. E. Green, Filip Rindler}

\address{$^1$The Alan Turing Institute \\
  London, United Kingdom\\
  e-mail: dgreen@turing.ac.uk \and
  $^2$
  Mathematics Institute, University of Warwick \\
  Coventry, United Kingdom\\
  e-mail: f.rindler@warwick.ac.uk}

\abstract{Model inference for dynamical systems aims to estimate the future behaviour of a system from observations. Purely model-free statistical methods, such as Artificial Neural Networks, tend to perform poorly for such tasks. They are therefore not well suited to many questions from applications, for example in Bayesian filtering and reliability estimation.

This work introduces a parametric polynomial kernel method that can be used for inferring the future behaviour of Ordinary Differential Equation models, including chaotic dynamical systems, from observations. Using numerical integration techniques, parametric representations of Ordinary Differential Equations can be learnt using Backpropagation and Stochastic Gradient Descent. The polynomial technique presented here is based on a nonparametric method, kernel ridge regression. However, the time complexity of nonparametric kernel ridge regression scales cubically with the number of training data points. Our parametric polynomial method avoids this manifestation of the curse of dimensionality, which becomes particularly relevant when working with large time series data sets.

Two numerical demonstrations are presented. First, a simple regression test case is used to illustrate the method and to compare the performance with standard Artificial Neural Network techniques. Second, a more substantial test case is the inference of a chaotic spatio-temporal dynamical system, the Lorenz--Emanuel system, from observations. Our method was able to successfully track the future behaviour of the system over time periods much larger than the training data sampling rate. Finally, some limitations of the method are presented, as well as proposed directions for future work to mitigate these limitations.
}

\begin{document}


\section{INTRODUCTION}

Dynamical systems play a crucial role in mathematical modelling across all areas of physics, engineering and applied mathematics. The equations used in some particular application domain are typically derived either phenomenologically \cite{clinton2010elegant} or from first principles such as the conservation of energy, mass or momentum (as in mechanics \cite{taylor2009}). The structure of the equations should describe the fundamental aspects of the system in question as much as possible. On the other hand, constitutive parameters are often hard to know explicitly and need to be learnt from data. As such, it is necessary to balance rigidity and flexibility when modelling a system.

This paper considers the problem of finding a model of a dynamical system, represented by coupled Ordinary Differential Equations (ODEs), from observations. This is a particular form of inverse problem (as in \cite{stuart2010}). The time evolution of many dynamical systems is described by polynomial equations in the system variables and their derivatives. We introduce a form of parametric polynomial kernel regression (related to Radial Basis Function networks \cite{aiModernApproach}). This technique was developed during the search for an algorithm that is able to be trained continuously on streaming data as opposed to complete trajectories. Hidden parameter models (with unobserved variables) are not addressed but the techniques shown here could be extended to such cases in the future, augmenting probabilistic Bayesian filtering methods (as in \cite{matthies2016}).

Kernel ridge regression is a nonparametric method for fitting polynomials to data without explicitly calculating all polynomial terms of a set of variables \cite{murphy2012machine,aiModernApproach}. There are two limitations of this approach when fitting models to time series data. First, as a nonparametric method, the computational time complexity scales cubically with the number of observation points. This is a significant issue when dealing with time series data. Second, it is difficult to compute kernel ridge regression efficiently using streaming data. While it is possible to continually update the inverse of a matrix (see \cite{Hager1989}), the roughly cubic scaling of the required matrix operations is not well suited to monitoring high-dimensional systems in a real time data setting. Here, to optimise our parametric polynomial kernel function representations, Stochastic Gradient Descent (SGD) is used along with the Backpropagation method (see \cite{bishop1995}). This combination of techniques helps to minimise computational complexity and the amount of explicit feature engineering required to find a good representation of an unknown ODE.


We represent ODE models parametrically as compute graphs. Compute graphs are used in Artificial Neural Network (ANN) theory to model complicated nonlinear structures by the composition of simple functions and are well suited to gradient descent optimisation via the Backpropagation method. It is demonstrated that numerical integration (both explicit and implicit) can be used to discretise ODE time integrals in a way that allows for the inference of continuous-time dynamical system models by gradient descent. This is an extension of an approach that appeared at least as early as \cite{Eberhard1999}. The discretisation procedure is related to the Backpropagation Through Time method \cite{werbos1988}, which is used for modelling discrete time series with so-called Recurrent Neural Networks.

To demonstrate the findings of this paper, two numerical case studies were carried out. The first is a simple analysis that contrasts the performance of standard ANN techniques with the proposed kernel method. It is shown that our method had the best extrapolation performance. A more extensive analysis of the chaotic spatio-temporal Lorenz--Emanuel dynamical system is also presented. The proposed method is able to recover a maximum likelihood estimate of the hidden polynomial model. For comparison, a parametric model constructed by direct summation of polynomial features (without kernels, of the form used in \cite{sudretHDR}) was also tested. The parametric polynomial kernel method was able to outperform the direct polynomial expansion, accurately predicting the future evolution of a chaotic dynamical system over periods many times greater than the training interval.

The primary advantage of the technique presented in this paper is that the model representation in parametric form can avoid the curse of dimensionality and poor scaling with training set size associated with nonparametric kernel regression. Further, polynomial kernels avoid the combinatorial explosion that occurs when explicitly computing polynomial series expansions. Interestingly, the accuracy of the proposed parametric kernel method can be tuned by adjusting the dimension of a set of intermediate parameters. The trade-off for increased accuracy is additional training time.

\section{BACKGROUND ON COMPUTE GRAPH OPTIMISATION}

\subsection{Compute graphs and nonlinear function representations}

The parametric polynomial regression technique introduced in this paper is built on the framework of so-called compute graphs. This section provides the background theory required for later parts of this work. Compute graphs are very general structures which define the flow of information over a topology and as such provide a convenient parametric representation of nonlinear functions. In particular, compute graphs can be coupled with Automatic Differentiation \cite{rall1981} and the Backpropagation algorithm (an application of the chain rule) to allow for gradient-based optimisation. Stochastic Gradient Descent is the most common form of optimiser used in this context and is briefly described in this section.

Artificial Neural Networks (ANNs) are a subset of compute graphs (in the sense of discrete mathematics \cite{eppDiscreteMath}). Common ANN terminology such as Deep Neural Networks, Boltzmann Machines, Convolutional Neural Networks and Multilayer Perceptrons refer to different ANN connectivity, training and subcomponent patterns \cite{bishop1995,Goodfellow-et-al-2016}. The choice of an appropriate ANN type depends on the problem being solved. This section works with general compute graph terminology, rather than specific ANN design patterns, as these principles are appropriate for all ANN architectures.

A (real-valued) compute graph consists of a weighted directed graph, i.e.\ an ordered pair $G = (V,E)$ with the following properties:
\begin{itemize}
  \item $V$ is the finite set of vertices (or nodes) $v_i$. Vertices specify an activation function $\sigma_i: \mathbb{R} \to \mathbb{R}$, and an output (or activation) value $a_i \in \mathbb{R}$.
  \item $E$ is the set of edges $e_{ij}$. Each edge $e_{ij}$ specifies a start vertex, defined to be $v_i$, and an end vertex, defined to be $v_j$. That is, edges are said to start at $v_i$ and terminate at $v_j$. Edges also specify a weight, $W_{ij} \in \mathbb{R}$.
\end{itemize}

\begin{figure}[t]
  \begin{center}
    \includegraphics[width=\textwidth]{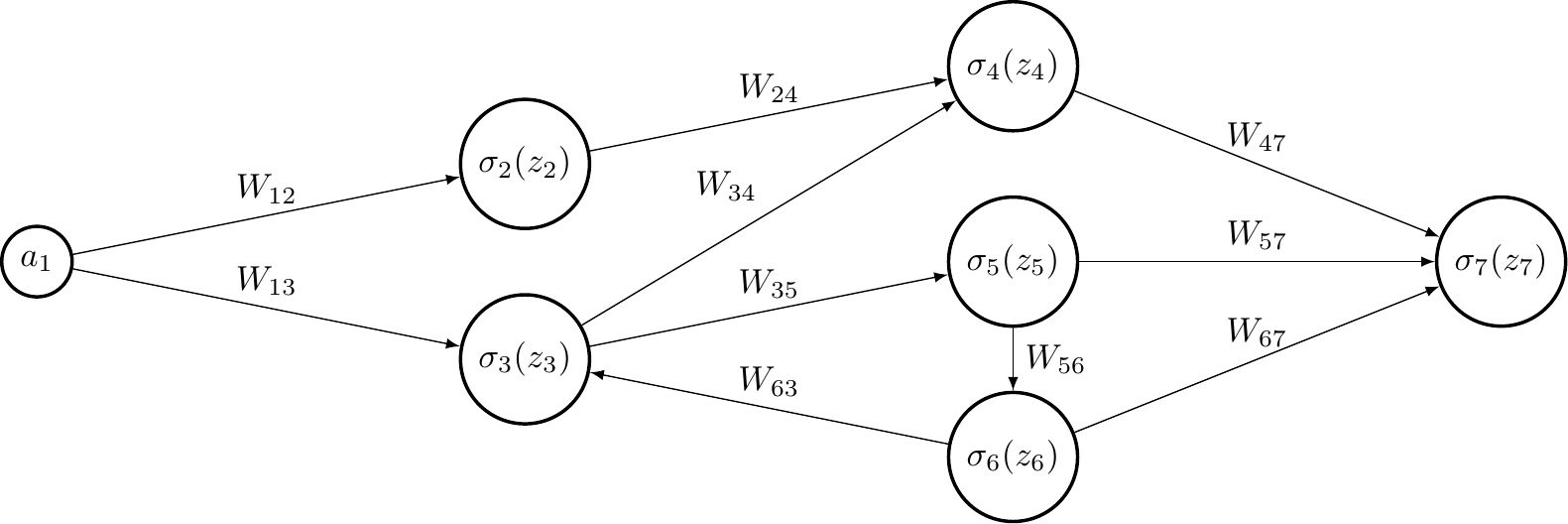}
    \caption{Example of compute graph. The subscript inside each node denotes the node number. Arrowheads indicate the direction of the graph edges. The function inside each node refers to the output function to be applied at the node. Note that node 1 is an input (with value $a_1$) as it has no parents. Further note that edge $W_{63}$ is recurrent as there is a cycle formed in the graph between nodes $3,5$ and $6$.}
    \label{fig:computeGraph}
  \end{center}
\end{figure}

Edges $e_{ij}$ can be understood as `pointing' from $v_i$ to $v_j$. Incoming edges to a node $v_i$ are all $e_{jk} \in E$ with $k = i$. Similarly, outgoing edges from a node $v_i$ are all $e_{jk} \in E$ with $j = i$. Parents of a node $v_i$ refer to all nodes $v_j$ such that there is an edge starting at $v_j$ and terminating at $v_i$. Similarly, children of a node $v_i$ refer to all nodes $v_j$ such that there is an edge starting at $v_i$ and terminating at $v_j$. A valid path of length $m$ starting at $v_1$ and terminating at $v_m$ is a set $\lbrace v_1, v_2, \cdots v_m \rbrace$ of at least two nodes such that there exist edges in $E$ from $v_i$ to $v_{i+1}$ for all $i \in [1,m-1]$. A recurrent edge in a compute graph refers to an edge that lies on a valid path from a node $v_i$ to any of its parents. A graph with recurrent edges is said to be a recurrent graph. An example of a (recurrent) compute graph is shown in fig \ref{fig:computeGraph}.

Inputs to the compute graph are all those nodes with no incoming edges (i.e.\ no parents), $\lbrace v_i | v_i \in V \wedge \nexists e_{ki} \in E \rbrace$. The activation values $a_i$ for input nodes $v_i$ must be assigned. The values at all other nodes, $v_i$, in the compute graph are calculated by
\begin{eqnarray}
  \label{eqn:zedI}
  z_i &=& \sum_{k:\text{ $v_k$ parent of $v_i$}} W_{ki} a_k, \\
  \label{eqn:activationI}
  a_i &=& \sigma_i \left( z_i \right),
\end{eqnarray}
where $z_i$ represents the weighted inputs to a node from all parent nodes and $a_i$ represents the output from a node.

Note that ANNs often define so-called bias units. Bias units allow for inputs to a node to have their mean easily shifted. A bias input to some node $v_i$ can be represented in a compute graph by creating a set of nodes $b_i \in B$, with no parents, that always output a value of $1$. Further, each $b_i$ is assigned to be an additional parent of $v_i$ by creating an edge from $b_i$ to $v_i$ with weight $B_i$ so that
\begin{eqnarray}
a_i = \sigma_i \left( \sum_{k:\text{ $v_k$ parent of $v_i$}} W_{ki} a_k + B_i \right).
\end{eqnarray}
Bias units will not, however, be explicitly indicated in the rest of this section as they can be assumed to be implicitly defined in eqn \eqref{eqn:zedI}.

The composition of simple functions with a compute graph structure allows for complicated nonlinear functions to be represented parametrically \cite{bishop1995}.

\subsection{Optimisation by Stochastic Gradient Descent and Backpropagation}

Optimisation over very large compute graphs representing highly nonlinear functions has become possible using Stochastic Gradient Descent (SGD) coupled with Backpropagation of errors \cite{bishop1995}. Advanced forms of SGD such as the Adam optimisation technique \cite{Kingma2014} are useful for optimising complicated compute graphs. The basic SGD method is described here. Stochastic Gradient Descent finds a locally optimal set of parameters, $\theta$, by iteratively updating the current estimate for the optimal parameters, $\theta_i$. It does so by moving the current estimate in the direction of greatest decreasing error, given by the derivative $\nabla_\theta J(\theta_i)$:
\begin{eqnarray}
  \label{eqn:gradDescentExpand}
    \theta_{i+1} := \theta_i - \eta \nabla_\theta J(\theta_i),
\end{eqnarray}
where $\eta$ is a small parameter that gives the distance to move in the direction defined by $\nabla_\theta J(\theta)$. Iterations are repeated until a specified error tolerance $\epsilon > 0$ is reached, i.e.\ until
\begin{eqnarray}
    J(\theta_i) \leq \epsilon.
\end{eqnarray}

Consider the case of approximating some unknown function $f(x)$ by a compute graph that outputs the function $\tilde{f}_\theta(x)$. The weights $\theta$ are taken to be the values of the edge weights $W_{ij}$ for all $e \in E$. Let the loss functional in this example be given by
\begin{eqnarray}
J(\theta) := \sum_{x} | f(x) - \tilde{f}_\theta(x) |^2,
\end{eqnarray}
for $x$ in some finite set. Thus, $J(\theta)$ is also representable as a compute graph. The graph for $J(\theta)$ contains the graph for $\tilde{f}_\theta(x)$ as a subset. To apply SGD to a compute graph, extended to contain the terms computing the loss functional, the Backpropagation method (an application of the chain rule) can be used if two conditions are met:
\begin{itemize}
  \item All nodal activation functions, $\sigma_i$, must be differentiable.
  \item The graph must be directed and acyclic, meaning the graph cannot contain any valid paths from a node to any of its parents, i.e. the graph must not have any recurrent edges.
\end{itemize}

If the above conditions are satisfied, Backpropagation can compute $\nabla_\theta J(\theta)$ via the chain rule. The basic procedure is outlined here, but a more detailed treatment can be found in \cite{bishop1995}. In the case that the graph is not acyclic, it can be unrolled via a technique referred to as Backpropagation Through Time \cite{werbos1988}.

Backwards error derivatives must be computed at all nodes, $v_i$, in the network:
\begin{eqnarray}
  \delta_{i} := \frac{\partial J}{\partial z_i}.
\end{eqnarray}
For nodes $v_i$ in the graph that compute the loss functional $J(\theta)$, the derivative $\delta_{i}$ can be computed directly. Otherwise, assume that node $v_i$ has children $\lbrace w_j \rbrace_{j=1}^N$. Using the chain rule, the error derivative $\delta_i$ can be calculated by pushing the error derivatives backwards through the graph from children to parents:
\begin{eqnarray}
  \delta_{i} = \sum_{j=1}^N \delta_j \frac{\partial z_j}{\partial a_i} \frac{\partial a_i}{\partial z_i} = \sum_{j=1}^N \delta_j W_{ij} \sigma_i' \left(z_i \right).
\end{eqnarray}
Given the error derivative terms, the desired error gradients $\nabla_\theta J(\theta)$ for $\theta = \lbrace W_{ij} \rbrace_{ij}$ can be computed at node $v_j$ with parents $\lbrace w_k \rbrace_{k=1}^M$ by
\begin{eqnarray}
\frac{\partial J}{\partial W_{ij}} = \delta_j \frac{\partial z_j}{\partial W_{ij}}
= \delta_j
\frac{\partial }{\partial W_{ij}} \left( \sum_{k=1}^M W_{kj}a_k \right) = \delta_j a_i.
\end{eqnarray}

Automatic Differentiation \cite{rall1981} can be used to write efficient computer code for Backpropagation. Specifically, Backpropagation is a form of `reverse accumulation mode' Automatic Differentiation. The above calculations can be organised efficiently by going through the compute graph from output to input nodes. At the time of writing, Tensorflow \cite{tensorflow2015-whitepaper} is a popular implementation of the algorithms described above. Although other (including gradient-free) optimisation procedures can be used that are suitable for general compute graphs, SGD with Backpropagation is typically very computationally efficient when applicable.

\section{PARAMETRIC POLYNOMIAL KERNEL REGRESSION}

\subsection{Overview}

Before discussing model inference for ODEs in particular, a parametric polynomial kernel function representation is introduced. Although ANNs and compute graphs are very effective at fitting arbitrary functions, standard ANN methods are poorly suited to polynomial function representation. As typical ANN architectures fit a very large number of parameters, they are unable to perform sensible extrapolation for even low-dimensional polynomial regression problems. Polynomial kernel ridge regression using the so-called kernel trick \cite{aiModernApproach} works well for fitting polynomials but suffers from cubic (that is, $\mathcal{O}(N^3)$) computational time complexity. Gradient-descent compute graph optimisation, as it is a parametric method, provides a way to optimise large data sets without the computational difficulties faced by nonparametric methods. While it is possible to build a compute graph that explicitly includes polynomial basis features, this scales factorially with the number of polynomial features included. In this paper it is shown that polynomial kernels can be inserted into compute graph structures and optimised by SGD, avoiding both the combinatorial explosion of polynomial series expansions and the poor time scaling of nonparametric kernel ridge regression.

\subsection{Polynomial kernel ridge regression}

Polynomial kernels, typically associated with kernel regression and Support Vector Machines \cite{aiModernApproach,murphy2012machine}, are functions of the form
\begin{eqnarray}
  \label{eqn:polyKernel}
  K(x,y) = \left( b \langle x,y\rangle + c\right)^d
\end{eqnarray}
for some $b,c \in \mathbb{R}$, $d \geq 1$. If the values of $y$ are assumed to be some parameters, the expansion of the polynomial kernel (for $d \in \mathbb{N}$) will, implicitly, yield all polynomial combinations up to order $d$.

Kernel ridge regression is a nonparametric method in the sense that the number of parameters grows with the amount of training data \cite{murphy2012machine}. By contrast, in this paper `parametric model' refers to a model with a fixed number of parameters. Adopting the notation in \cite{vu2015}, the standard form of ridge regression is as follows. Given observations of an unknown function $f \colon \mathbb{R}^D \to \mathbb{R}^E$ at $N$ locations, $\lbrace (x_i, f(x_i))\rbrace_{i=1}^N$, kernel ridge regression finds an approximation, $f_k(x)$, by
\begin{eqnarray}
  \label{eqn:kernelRidgeRegression}
  f(x) \approx f_k(x) = \sum_{i=1}^N \alpha_i K(x,x_i),
\end{eqnarray}
where the values $\alpha_i$ are termed weights and $K(x,x_i)$ is a kernel function. Kernel functions are a form of generalisation of positive definite matrices (see \cite{murphy2012machine} for additional details). Only the (real-valued) polynomial kernel in eqn \eqref{eqn:polyKernel} will be discussed in this paper. The weights $\alpha = \left( \alpha_1, \ldots, \alpha_N \right)$ are calculated using $f(x) = \left( f(x_1), \ldots, f(x_N) \right)$ as follows:
\begin{eqnarray}
  \label{eqn:kernelRidgeRegressionAlpha}
  \alpha = \left(K + \lambda I \right)^{-1} f(x),
\end{eqnarray}
where $K \in \mathbb{R}^{N \times N}$ is the matrix with entries $K_{ji} = K(x_j,x_i)$ and $I$ is the $N$ by $N$ identity matrix. The term $\lambda \in \mathbb{R}$ is a regularisation term that controls overfitting. Note that if $K + \lambda I$ is not invertible, then the inverse must be replaced by a pseudo-inverse. In the sense of Bayesian regression, the term $\lambda$ represents the scale of Gaussian noise added to observations $f(x_i)$ as a part of the approximation procedure.

The use of kernels for regression as in eqn \eqref{eqn:kernelRidgeRegression} has the effect of mapping a low-dimensional problem implicitly into a high-dimensional space. This is a very powerful technique for projecting data onto high-dimensional basis functions. Unfortunately, as a (typically) dense matrix must be inverted to calculate $\alpha$, the computational complexity of standard kernel ridge regression scales cubically with the number of data points, $N$. This is a severe limitation when considering large data sets such as the time series data considered in later sections of this paper.

\subsection{Parametric polynomial kernel representation}

Instead of calculating an inner product between known values of $x$ and $y$ as in eqn \eqref{eqn:polyKernel} and inverting a matrix as in eqn \eqref{eqn:kernelRidgeRegressionAlpha}, this paper demonstrates that a kernel representation can be found in an efficient way using compute graphs and SGD. Consider the following parametric representation of a function $f \colon \mathbb{R}^D \to \mathbb{R}^E$ with parameters $\theta \in \Theta$:
\begin{eqnarray}
  \label{eqn:secondOrderPolyModel}
  f_\theta(x) = W_2\left[ (W_1x + B_1) \circ (W_1x + B_1)  \right] + B_2,
\end{eqnarray}
where $\circ$ denotes elementwise matrix multiplication (or Hadamard product), i.e.\ $A = B \circ C$ means $a_{ij} = b_{ij}c_{ij}$ for the corresponding matrix entries \cite{hornJohnson2012}. The remaining terms are defined by $W_1 \in \mathbb{R}^{M\times D}$, $B_1 \in \mathbb{R}^{D}$, $W_2 \in \mathbb{R}^{E\times M}$ and $B_2 \in \mathbb{R}^{E}$. The parameters $B_1, B_2$ are known as bias weights in the ANN literature \cite{bishop1995}. The full set of parameters for this model is $\theta = \lbrace W_1, B_1, W_2, B_2 \rbrace$. The dimension $M$ is an intermediate representation dimension and is discussed below.

Eqn \eqref{eqn:secondOrderPolyModel} is a parametric representation of a second-order polynomial kernel. Expanding eqn \eqref{eqn:secondOrderPolyModel} explicitly would yield a set of second-order polynomials in terms of $x_i$. However, using SGD the unknown polynomial expression can be found without the need to know the expanded polynomial form. The elementwise matrix product acts like the $d$-th power in eqn \eqref{eqn:polyKernel}. The parameters $\theta$ can be trained by SGD and function as parametric representations of Support Vectors. The term $M$ required to complete the definition of eqn \eqref{eqn:secondOrderPolyModel} is a hyperparameter representing a choice of intermediate representation dimension and is related to the number of Support Vectors required to represent the system (as in Support Vector Regression, see \cite{aiModernApproach}). Increasing the size of $M$ increases the number of parameters but can improve the fit of the regressor (as is demonstrated empirically in Section \ref{sec:mainNumerical}).

An $n$-th order polynomial could be fit by taking a larger number of Hadamard products. Denote the composition of Hadamard products by $A \circ^n A := A \circ A \circ \cdots \circ A$ ($n$ times). Then, our approach consists of expressing an $n$-th order representation of $f_\theta \colon \mathbb{R}^D \to \mathbb{R}^E$ as follows:
\begin{eqnarray}
  \label{eqn:nOrderPolyModel}
  f_\theta(x) = W_2\left[ (W_1X + B_1) \circ^{n} (W_1 X + B_1)\right] + B_2
\end{eqnarray}
or some similar variation on this theme. The expression in eqn \eqref{eqn:nOrderPolyModel} is differentiable in the sense of compute graphs since all of the operations in eqn \eqref{eqn:nOrderPolyModel} are differentiable. Comparing with eqns \eqref{eqn:kernelRidgeRegression} and \eqref{eqn:kernelRidgeRegressionAlpha}, the parametric form of polynomial kernel regression can be thought of as an approximation to both the $\alpha_i$ and $K(x,x_i)$ terms in a single expression. As the parametric regression form can be optimised by SGD, the cubic scaling of nonparametric kernel ridge regression is avoided.

\subsection{Numerical demonstration on simple regression problem}

This section demonstrates the proposed method via the approximation of a simple cubic function, namely
\begin{eqnarray}
\label{eqn:trueFuncPolyTest}
f(x) := (x-1)(x+1)(x+0.5).
\end{eqnarray}

The goal of this analysis is to infer the hidden function $f(x)$. Given a set of training data, $N$ pairs $\lbrace(x_i,f(x_i))\rbrace_{i=1}^N$, the problem is to minimise the loss functional
\begin{eqnarray}
  \label{eqn:simpleRegressionLoss}
J(\theta) := \frac{1}{N} \sum_{i=1}^N | f(x_i) - f_\theta(x) |^2.
\end{eqnarray}
For this test problem, $N=25$ training data points were sampled uniformly between $x=-2$ and $x=2$.

First, a standard ANN `Multilayer Perceptron' (specifically a three-layer deep, 100 unit wide perceptron network) was tested. The reader unfamiliar with these terms can see \cite{aiModernApproach} for definitions, but it is sufficient for the purposes of this paper to understand that this perceptron model computes the function
\begin{eqnarray}
\label{eqn:multilayerpercep}
f_\theta(x) = W_4\sigma(W_3 \sigma(W_2 \sigma(W_1 x + B_1) + B_ 2) + B_3) + B_4
\end{eqnarray}
where $W_1 \in \mathbb{R}^{100 \times 1}$, $W_2,W_3 \in \mathbb{R}^{100 \times 100}$, $W_4 \in \mathbb{R}^{1 \times 100}$, $B_1,B_2,B_3 \in \mathbb{R}^{100}$, and $B_4 \in \mathbb{R}$ such that the parameters of this network are $\theta = \lbrace W_i,B_i \rbrace_{i=1}^4$. Additionally, $\sigma(x)$ denotes the sigmoid function:
\begin{eqnarray}
  \sigma(x) := \frac {1}{1+e^{-x}}.
\end{eqnarray}
In eqn \eqref{eqn:multilayerpercep}, $\sigma$ is applied to vectors componentwise.

Second, the parametric polynomial method in eqn \eqref{eqn:nOrderPolyModel} was tested for polynomial orders $n=2,3,4$. The parameter $M$ was fixed to $20$ for all comparisons.

Both the perceptron model and the parametric polynomial kernel model were trained in two stages. The Adam optimiser \cite{Kingma2014} was first run for 1000 iterations with a learning rate of $0.01$ and then for an additional 1000 iterations with a learning rate of $0.001$. All ANNs and SGD optimisers were implemented using the Tensorflow software library \cite{tensorflow2015-whitepaper}.

Finally, a nonparametric kernel ridge regression estimator of the form in eqn \eqref{eqn:kernelRidgeRegression} was tested. This was implemented using the SciKit learn `KernelRidge' function \cite{scikit2011} using a third-order polynomial kernel. Note that this function has additional hyperparameters, $\alpha, \mathrm{coef0}$ and $\gamma$. These were set to $0.1$, $10$ and `None' respectively. The SciKit documentation describes these parameters in detail. As with the parametric estimator, the choice of maximum polynomial degree ($d$ in eqn \eqref{eqn:kernelRidgeRegressionAlpha}) is another hyperparameter. For this demonstration, only the known true value ($d=3$) was tested with the nonparametric regression estimator.

\begin{table}[ht]
  \begin{center}
    \begin{tabular}{c l}
    \hline
    Function representation & $J(\theta)$ \\
    \hline
    Multilayer Perceptron & $1.12 \times 10^{-4}$ \\
    Parametric kernel with $n=2$ & $1.70 \times 10^{0}$ \\
    Parametric kernel with $n=3$ & $5.95 \times 10^{-14}$ \\
    Parametric kernel with $n=4$ & $2.34 \times 10^{-1}$ \\
    Nonparametric polynomial kernel & $9.68 \times 10^{-3}$ \\
    \hline
    \end{tabular}
  \end{center}
\caption{Values of $J(\theta)$, defined in eqn \eqref{eqn:simpleRegressionLoss}, after optimisation by SGD for the simple regression task. \label{tab:simpleRegressionJ}}
\end{table}

The values of $J(\theta)$ after running SGD are shown in table \ref{tab:simpleRegressionJ}. The third-order parametric polynomial loss is ten orders of magnitude lower than the regression loss of the perceptron network. The lower loss of the $n=3$ parametric polynomial method compared to $n=2$ and $n=4$ is (of course) expected as the hidden function is a third-order polynomial. This indicates that several polynomial orders should be tested when applying the proposed technique to other problems.

\begin{figure}[tb]
  \begin{center}
    \includegraphics{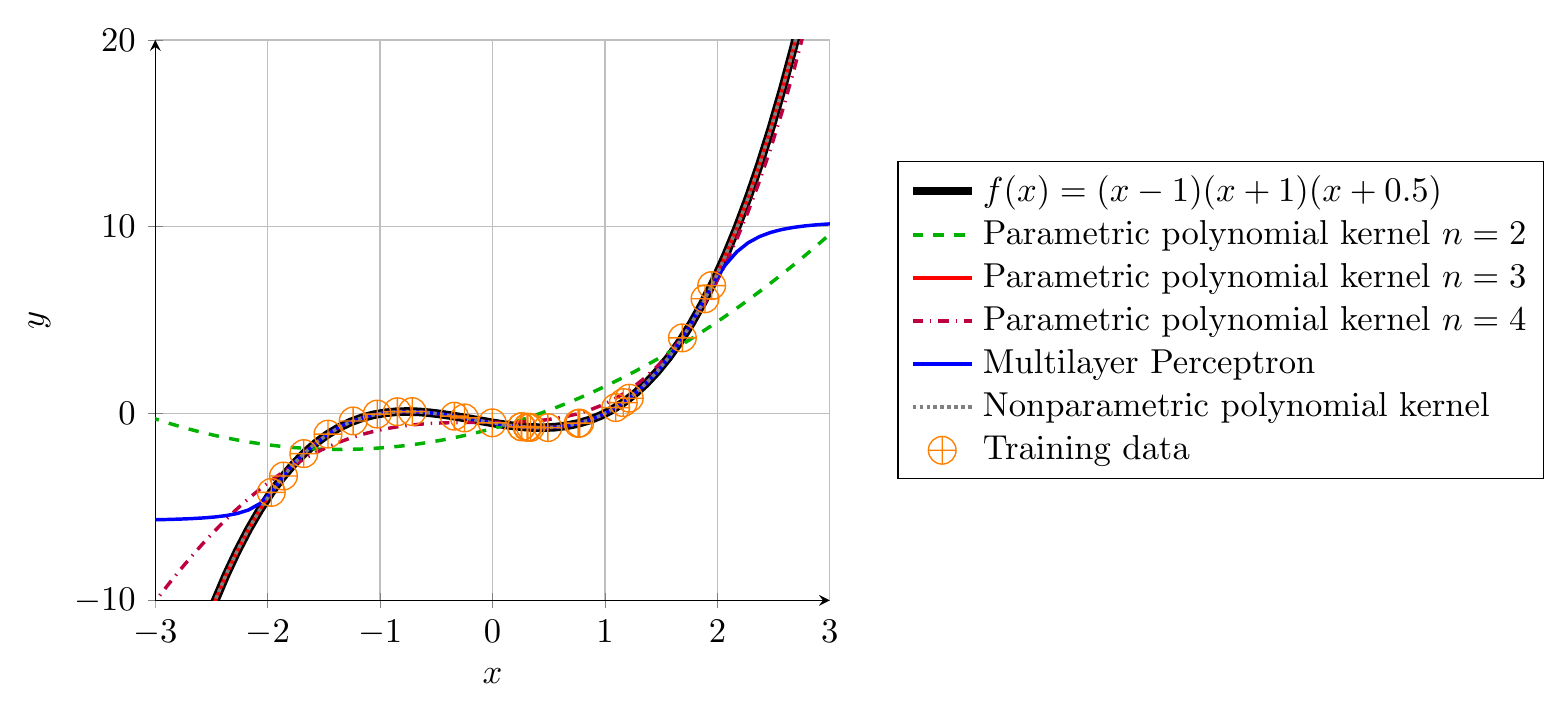}
    \caption{Comparison of performance of the parametric polynomial kernel method on a simple regression task. Note that the true hidden function, from eqn \eqref{eqn:trueFuncPolyTest}, is underneath the function inferred by the $n=3$ parametric polynomial. The two coincide because of the virtually perfect fit. The nonparametric polynomial kernel ridge estimator also closely coincides with the true $f(x)$. The 25 regression training data points were calculated by sampling uniformly between $x=-2$ and $x=2$.}
    \label{fig:fittedPolynomialTest}
  \end{center}
\end{figure}

\begin{figure}[tb]
  \begin{center}
    \includegraphics{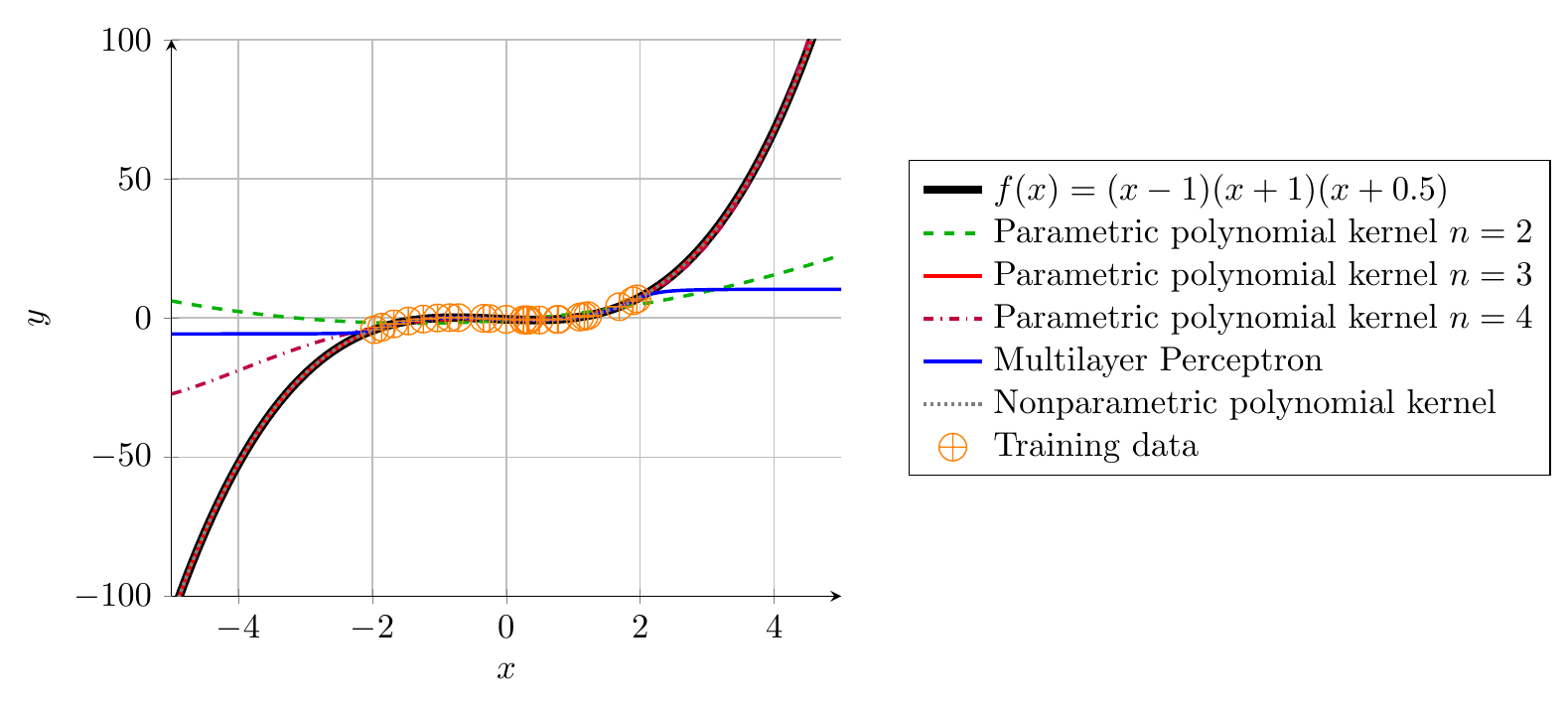}
    \caption{Comparison of performance of the parametric polynomial kernel method on a simple regression task. This is a zoomed out view of fig \ref{fig:fittedPolynomialTest} and shows that the polynomial kernel estimators (both parametric for $n=3$ and nonparametric) are able to recover the true hidden function in eqn \eqref{eqn:trueFuncPolyTest} outside of the range of the training data.}
    \label{fig:fittedPolynomialTestZoomOut}
  \end{center}
\end{figure}

\begin{figure}[tb]
  \begin{center}
    \includegraphics{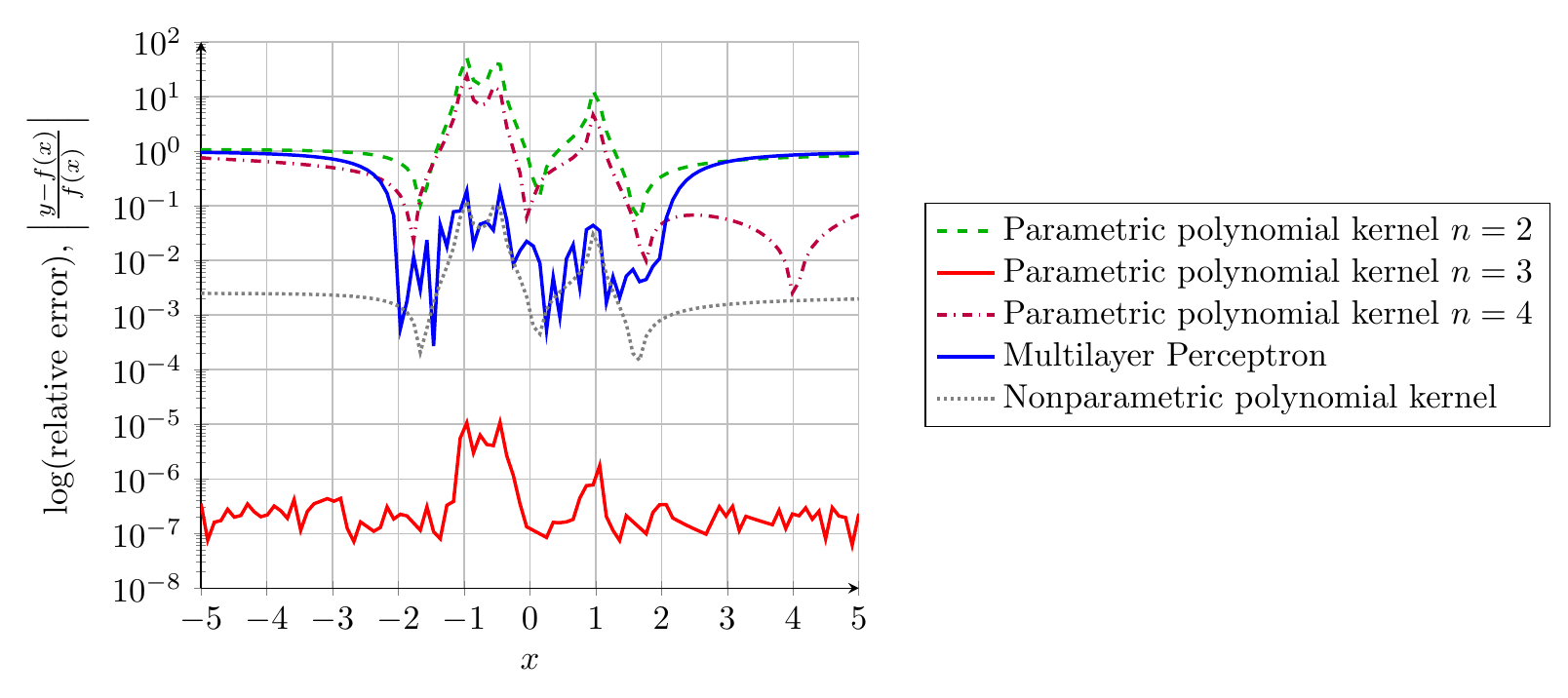}
    \caption{Comparison of pointwise absolute errors for the simple regression task. Errors are computed as $\left\lvert \frac{y-f(x)}{f(x)} \right\rvert$ where $f(x)$ is the true hidden function defined in eqn \eqref{eqn:trueFuncPolyTest}. The parametric polynomial kernel method has the best performance, followed by the nonparametric polynomial kernel ridge method. Note that the training data was restricted to lie within $x=-2$ and $x=2$.}
    \label{fig:fittedPolynomialTestError}
  \end{center}
\end{figure}

The results of the analysis are shown in figs \ref{fig:fittedPolynomialTest} and \ref{fig:fittedPolynomialTestZoomOut}. Each model tested was able to recover the true form of $f(x)$ in the region of the training data. Relative errors for each method are shown in fig \ref{fig:fittedPolynomialTestError}. Both the parametric and nonparametric polynomial methods were also able to extrapolate well beyond the range of the original data for the $n=3$ model. This can be best seen in fig \ref{fig:fittedPolynomialTestZoomOut}. The perceptron model, by contrast, almost immediately fails to predict values of the hidden function outside of range of the training data. For inferring hidden polynomial dynamical systems from observations, where the ability to extrapolate beyond the training data is essential, the analysis in this section suggests that the parametric polynomial kernel method can be expected to have performance superior to standard ANN methods.

This analysis also indicates that the loss $J(\theta)$ is an effective indicator of extrapolation performance for polynomial kernel methods (at least in this test case). This is not true for the Multilayer Perceptron model which had a low $J(\theta)$ value but poor extrapolation performance. One must however take care when making assertions about extrapolation performance, as it is easy to make incorrect inferences in the absence of data.

\clearpage

\section{ORDINARY DIFFERENTIAL EQUATION MODEL INFERENCE}

\subsection{Dynamical Systems}

Dynamical systems are classified into either difference equations (discrete-time systems) or differential equations (continuous-time systems) \cite{meiss2017}. In this paper, only continuous-time dynamical systems are investigated, although the numerical methods presented could be applied to both continuous-time and discrete-time systems. Continuous-time dynamical systems of the form considered in this paper can be expressed as coupled first-order Ordinary Differential Equations (ODEs):
\begin{eqnarray}
  \label{eqn:odeForm}
  \frac{d}{dt}u(t) = f(t,u(t)),
\end{eqnarray}
where:
\begin{itemize}
  \item $t \in [0,\infty)$ represents time;
  \item $u(t) \in \mathbb{R}^n$ is the vector of values representing the $n$ variables of the system at time $t$;
  \item $f(t,u(t)) \in \mathbb{R}^n$ represents the prescribed time derivatives of $u(t)$.
\end{itemize}

A trajectory of a dynamical system refers to a parameterised path $u(t)$ which returns a value of $u$ for all values of the parameter $t$. The value of $u(t)$ in eqn \eqref{eqn:odeForm} can be computed given some initial value, $u(0)$, by integrating $f(t,u(t))$ forward in time:
\begin{eqnarray}
  \label{eqn:odeIntegralForm}
  u(t) = u(0) + \int_0^t \frac{d}{d\tau}u(\tau) d\tau = u(0) + \int_0^t f(\tau,u(\tau)) d\tau
\end{eqnarray}

To simplify the solution of ODEs and the implementation of the learning algorithm presented in this paper, we only consider first-order systems. A differential equation of order $m$ of the form
\begin{eqnarray}
  \frac{d^m}{dt^m}u(t) = f(t,u(t))
\end{eqnarray}
can be converted into a system of first-order coupled ODEs. This is also the standard approach employed in numerical implementations of ODE solvers, for an example, see the SciPy function solve$\_$ivp \cite{scipy2001}. The conversion can be achieved by introducing new variables for higher derivatives. Consider an $m$-th order equation of the form
\begin{eqnarray}
  \frac{d^{m}u}{dt^{m}}  = g\left(t,u,\frac{du}{dt},\frac{d^2u}{dt^2}, \cdots, \frac{d^{m-1}u}{dt^{m-1}} \right).
\end{eqnarray}

This can be rewritten by replacing the $\frac{d^{i}u}{dt^{i}}$ terms by new variables $v_i$ ($i \in [1,m-1]$) such that:
\begin{eqnarray}
   \frac{d}{dt}
   \begin{bmatrix}
     u \\ v_1 \\ \vdots \\ v_{m-1}
   \end{bmatrix}
   =
   \begin{bmatrix}
     v_1 \\ \vdots \\ v_{m-1} \\ f(t,u,v_1,v_2, \dots, v_{m-1})
   \end{bmatrix}.
\end{eqnarray}

As the value of $u$ at some time depends on the values at infinitesimally earlier times through the derivatives of $u$, there is a recursive structure present in the equations (this would be even clearer for difference equations or after a discretisation). The model inference technique presented in this paper uses loop unrolling to simplify the derived optimisation problem.

\subsection{Model inference for coupled ODEs}

Model inference, in this context, is the problem of recovering the form of $f(t,u(t))$ (as in eqn \eqref{eqn:odeForm}) given observations of $u(t)$ at times from $0$ to $T$. Model inference can be expressed as an optimisation problem:
\begin{eqnarray}
  \label{eqn:integralOpt}
  \operatorname{Minimise}\; J(\theta) := \int_0^T \biggm\lvert u(t) - \left( u(0) + \int_0^t f_\theta(\tau,u(\tau)) d\tau \right) \biggm\lvert^2 dt,
\end{eqnarray}
where $J(\theta)$ is a loss functional over some unknown parameters $\theta \in \Theta$. The function $f_\theta(\tau,u(\tau))$ denotes a parametric approximation to the true latent function $f(t,u(t))$. For the purposes of this paper, the parametric representation of $f_\theta$ can be assumed to be a directed acyclic compute graph. Denote the trajectories computed using the integral of $f_\theta$ by
\begin{eqnarray}
  \label{eqn:fthetaApproxTrajectory}
  \tilde{u}_\theta(t) := u(0) + \int_0^{t} f_\theta(\tau,u(\tau)) d\tau.
\end{eqnarray}
Then the loss functional in eqn \eqref{eqn:integralOpt} can be expressed as
\begin{eqnarray}
  \label{eqn:integralOptTrajectory}
  J(\theta) = \int_0^T \left\lvert u(t) - \tilde{u}_\theta(t) \right\rvert^2 dt.
\end{eqnarray}
In this form, it is clear that the $J(\theta)$ measures how closely the observed trajectories $u(t)$ match the predicted trajectories $\tilde{u}_\theta(t)$ for each value of $\theta$. Additionally, although the $\mathrm{L}^2$ norm has been used above, this norm could be changed to any other norm as appropriate. For simplicity, only the $\mathrm{L}^2$ norm will be used in this paper.

If observations of $\frac{du}{dt}$ are available, the optimisation problem can be expressed in an alternative, but not exactly equivalent, differential form:
\begin{eqnarray}
  \label{eqn:diffOpt}
  \operatorname{Minimise}\; K(\theta) := \int_0^T \biggm\lvert \frac{d}{dt}u(t) - f_\theta(t,u(t)) \biggm\lvert^2 dt.
\end{eqnarray}

The loss functional surface for $J(\theta)$ will tend to be smoother over $\theta$ when compared to the differential form (since there is an additional integration), potentially altering the behaviour of various optimisation methods. However, the exact minimisers $\theta^\ast$ of both $J(\theta)$ and $K(\theta)$, if they exist so that $J(\theta^\ast) = K(\theta^\ast) = 0$, are the same, as can be seen by differentiation.

The choice to optimise over $K(\theta)$ or $J(\theta)$ depends on the chosen representation of $f_\theta$ and the availability of observations. Assume that only observations of $u(t)$ are available and not direct observations of $\frac{du}{dt}$. Then it is necessary to either introduce some way to approximate $\frac{du}{dt}$ or to approximate $\int_0^t f_\theta(\tau,u(\tau)) d\tau$. In the remainder of this section, it is shown that a discretised form of $J(\theta)$, denoted by $\hat{J}(\theta)$, can be derived. The discretised objective $\hat{J}(\theta)$ can be trained using SGD and Backpropagation as long as $f_\theta (t,u(t)) $ can be represented by an acyclic compute graph. The derivation of $\hat{J}(\theta)$ proceeds by first approximating the outer integral in eqn \eqref{eqn:integralOptTrajectory} using a finite set of observations of $u(t)$. The derivation of the discretisation is completed by approximating the integral $\int_0^t f_\theta (\tau,u(\tau)) d \tau$ using standard numerical time integration techniques.

\subsection{Discretisation of the approximate trajectories}

The continuous form of the integral in eqn \eqref{eqn:fthetaApproxTrajectory} is not amenable to numerical computation and requires discretisation. In particular, if $f_\theta$ is to be represented by a compute graph and learnt by SGD, then the entire loss functional $J(\theta)$ must be represented by a differentiable, directed acyclic compute graph. To achieve this, it is useful to first note that the integral in eqn \eqref{eqn:fthetaApproxTrajectory} can be decomposed into a series of integrals over smaller time domains. Consider the trajectories from times $0$ to $t$ and $0$ to $t+h$:
\begin{eqnarray}
  \tilde{u}_\theta(t) := u(0) + \int_{0}^{t} f_\theta(\tau,u(\tau)) d\tau.
\end{eqnarray}
Then,
\begin{eqnarray}
  \tilde{u}_\theta(t+h)
  &=& u(0) + \int_{0}^{t} f_\theta(\tau,u(\tau)) d\tau + \int_{t}^{t+h} f_\theta(\tau,u(\tau)) d\tau\\
  \label{eqn:trajectoryDiscretisationOne}
  &=& \tilde{u}_\theta(t) + \int_{t}^{t+h} f_\theta(\tau,u(\tau)) d\tau,
\end{eqnarray}
giving the trajectory predicted by $f_\theta$ from $\tilde{u}_\theta(t)$ to $\tilde{u}_\theta(t+h)$.

The required discretisation can be completed using standard numerical integration techniques. Numerical integration methods such as Euler, Runge-Kutta and Backwards Differentiation (see \cite{iserles2009} for an overview) work, roughly, by assuming some functional form for $f(x)$ and analytically integrating this approximation. Numerical integration methods can be expressed as a function of the integrand evaluated at some finite set of $m$ points $\lbrace x_j \rbrace_{j=1}^m$:
\begin{eqnarray}
  \label{eqn:numericalApprox}
    \int_{a}^{b} f(x) dx \approx G\left(a,b,f,\lbrace x_j \rbrace_{j=1}^m \right).
\end{eqnarray}
Note that the points $a \leq x_j \leq b$ are defined as a part of the specification of a particular numerical integration scheme. The function to be integrated, $f$, must be able to be evaluated at each $x_j$.

The trajectories in eqn \eqref{eqn:trajectoryDiscretisationOne} can then be approximated with a numerical approximation scheme as in eqn \eqref{eqn:numericalApprox}:
\begin{eqnarray}
  \label{eqn:trajectoryDiscretisation}
  \tilde{u}_\theta(t+h) \approx \hat{u}_\theta(t+h) &:=& \hat{u}_\theta(t_j) + G\left(t,t+h,f_\theta,\lbrace (\tau_j,u(\tau_j)) \rbrace_{j=1}^m \right), \\
  \hat{u}_\theta(0) &:=& u(0).
\end{eqnarray}
$\hat{u}_\theta(t)$ refers to a trajectory $\tilde{u}_\theta(t)$ with continuous integrals replaced by approximate numerical integrals. The values $\tau_j$ are evaluation points and correspond to the values $x_j$ in eqn \eqref{eqn:numericalApprox}. In general, the smaller the value of $h$ the greater the accuracy of the approximation. Small values of $h$, however, increase the computational burden required to compute approximate trajectories.

\subsection{ODE inference loss functional for observations at discrete times}

For practical problems, observations of $u(t)$ will not be available for all times between $0$ and $T$. Typically, the trajectory $u(t)$ will be known only at a finite set of times $t \in \lbrace t_i \rbrace_{i=1}^N$ so that $u(t)$ is known at $\lbrace u(t_i) \rbrace_{i=1}^N$. The finite set $\lbrace (t_i,u(t_i)) \rbrace_{i=1}^N$ will be referred to as `training data' and can be used to discretise the optimisation problem in eqn \eqref{eqn:integralOpt} by the following approximation:
\begin{eqnarray}
  \label{eqn:integralOptDiscreteOne}
  \operatorname{Minimise}\; \tilde{J}(\theta) &:=& \frac{1}{N}\sum_{i=1}^N \left\lvert u(t_i) - \left( u(0) + \int_0^{t_i} f_\theta(\tau,u(\tau)) d\tau \right) \right\lvert^2\\
  \label{eqn:integralOptDiscreteOneOne}
  &=& \frac{1}{N}\sum_{i=1}^N \bigl\lvert u(t_i) - \tilde{u}_\theta(t_i) \bigr\lvert^2.
\end{eqnarray}

However, the terms $\tilde{u}_\theta(t)$ must also be replaced by a discretisation, as in eqn \eqref{eqn:trajectoryDiscretisation}. Assume that a numerical integration scheme is selected that evaluates the integrand at $m$ points. It is convenient to decompose the trajectory integrals $\tilde{u}_\theta(t)$ into a series of integrals over finite subsets of the training data, $t_i$ to $t_{i+p}$ for the window size $p \in \mathbb{N}$ (typically either $m$ or $m-1$), such that
\begin{eqnarray}
  \tilde{u}_\theta(t_{i+p}) = \tilde{u}_\theta(t_i) + \int_{t_i}^{t_{i+p}} f_\theta(\tau,u_\theta(\tau)) d\tau.
\end{eqnarray}
With reference to eqn \eqref{eqn:trajectoryDiscretisation}, this can be further approximated by numerical integration:
\begin{eqnarray}
  \label{eqn:partiallySimplifiedUHat}
  \hat{u}_\theta(t_{i+p}) = \hat{u}_\theta(t_i) + G\left(t_i,t_{i+p},f_\theta,\lbrace (\tau_j,u(\tau_j)) \rbrace_{j=1}^m\right)
\end{eqnarray}
such that the value of $u(\tau_j)$ is known (given the training data) for all evaluation points $\tau_j$, $j \in [1,m]$.

Finally, eqn \eqref{eqn:partiallySimplifiedUHat} can be modified by using the known value (from the training data) of $u(t_i)$ in place of $\hat{u}_\theta(t_i)$:
\begin{eqnarray}
  \label{eqn:simplifiedUHat}
  \hat{u}(t_{i+p}) := u(t_i) + G\left(t_i,t_{i+p},f_\theta,\lbrace (\tau_j,u(\tau_j)) \rbrace_{j=1}^m \right).
\end{eqnarray}

Eqn \eqref{eqn:integralOptDiscreteOneOne} can be approximated by the discretised loss functional $\hat{J}(\theta)$ by inserting $\hat{u}(t)$:
\begin{eqnarray}
  \label{eqn:integralOptDiscrete}
  \hat{J}(\theta) &:=& \frac{1}{N-p}\sum_{i=1}^{N-p} \bigl\lvert u(t_{i+p}) - \hat{u}(t_{i+p}) \bigr\lvert^2\\
  \label{eqn:integralOptDiscreteFinal}
  &=& \frac{1}{N-p}\sum_{i=1}^{N-p} \bigl\lvert u(t_{i+p}) - \left( u(t_i) + G\left(t_i,t_{i+p},f_\theta,\lbrace (\tau_j,u(\tau_j)) \rbrace_{j=1}^m \right) \right) \bigr\lvert^2.
\end{eqnarray}
As $\hat{J}(\theta)$ is a discrete approximation to $J(\theta)$, the model inference problem in eqn \eqref{eqn:integralOptTrajectory} is approximately solved by minimisation of $\hat{J}(\theta)$ over a training data set:
\begin{eqnarray}
  \theta^\ast = \operatorname{argmin}_{\theta} J(\theta) \approx \operatorname{argmin}_{\theta} \hat{J}(\theta).
\end{eqnarray}
The inferred ODE model then is $f_{\theta^\ast}(t,u(t))$.

Note that in the above derivation, loss functionals have been computed for time-dependent models of the form $f(t,u(t))$. In practice, optimisation over a single trajectory will only provide useful estimates of $f_\theta$ very close to $(t,u(t))$. To find estimates of $f_\theta$ away from those points, one would have to observe multiple trajectories and modify $J(\theta)$ to average over these trajectories. Alternatively, in the autonomous case, where $f$ is of the form $f(u(t))$, one trajectory may be enough to infer $f_\theta$, depending on the number of sampling points available.


\subsection{Example using Euler integration}

To demonstrate concretely how eqn \eqref{eqn:integralOptDiscreteFinal} gives a loss functional discretisation, $\hat{J}(\theta)$, for an ODE model that can be optimised by SGD and Backpropagation, an example using simple numerical integration techniques is discussed in this section. Forward Euler (see \cite{iserles2009}) computes an approximation to a dynamical system trajectory time integral as follows ($h > 0$):
\begin{eqnarray}
  \label{eqn:forwardEuler}
  u(t+h) \approx u(t) + h f(t,u(t)).
\end{eqnarray}

With reference to eqn \eqref{eqn:numericalApprox}, Forward Euler is a numerical integration scheme with $m = p = 1$, $\tau_{1} = a$ and
\begin{eqnarray}
  \label{eqn:forwardEulerG}
  G\left(a,b,f,\lbrace (a,u(a)) \rbrace \right) = |b-a| f(a,u(a)).
\end{eqnarray}

Forward Euler is a so-called explicit method as the approximation of $u(t+h)$ depends only on functions evaluated at times earlier than $t+h$. Backward Euler, conversely, is an implicit method:
\begin{eqnarray}
  \label{eqn:backwardEuler}
  u(t+h) \approx u(t) + h f(t+h,u(t+h)).
\end{eqnarray}
With reference to eqn \eqref{eqn:numericalApprox}, Backward Euler is a numerical integration scheme with $m = p = 1$, $\tau_1 = b$, and
\begin{eqnarray}
  \label{eqn:backwardEulerG}
  G\left(a,b,f,\lbrace (b,u(b)) \rbrace \right) = |b-a| f(b,u(b)).
\end{eqnarray}

Forward time integration using Backward Euler requires solving a system of equations (typically by Newton-Raphson iterations \cite{iserles2009}) as $u(t+h)$ appears on both sides of eqn \eqref{eqn:backwardEuler}. This is characteristic of implicit integration methods. The choice of when to use explicit or implicit integration methods for simulation of a system depends on the form of the dynamical system to be approximated \cite{iserles2009}. Implicit methods are more efficient and accurate for so-called `stiff' problems \cite{Hairer1993,Hairer1996}.

However, either method can be used to discretise an ODE into a compute graph representation. For example, assume that $f_\theta(t,u(t))$ is represented by an acyclic compute graph. Then, given training data $\lbrace(t_i,u(t_i))\rbrace_{i=1}^N$, the model inference loss functional, $\hat{J}(\theta)$, in eqn \eqref{eqn:integralOptDiscreteFinal} can be approximated using Forward Euler as follows:
\begin{eqnarray}
  \label{eqn:forwardEulerBPTODE}
  \hat{J}_F(\theta) := \frac{1}{N-1} \sum_{i=1}^{N-1} \bigl\lvert u(t_{i+1}) - \bigl( u(t_i) + |t_{i+1}-t_{i}| f_\theta(t_i,u(t_i)) \bigr) \bigr\lvert^2.
\end{eqnarray}

Implicit integration schemes can be used in essentially the same way as shown above for Forward Euler. As an example, the Backward Euler scheme in \eqref{eqn:backwardEuler} can be used to set the model inference loss functional, $\hat{J}(\theta)$, from eqn \eqref{eqn:integralOptDiscreteFinal} as follows:
\begin{eqnarray}
  \label{eqn:backwardEulerBPTODE}
  \hat{J}_B(\theta) := \frac{1}{N-1} \sum_{i=1}^{N-1} \bigl\lvert u(t_{i+1}) - \bigl( u(t_i) + |t_{i+1}-t_{i}| f_\theta(t_{i+1},u(t_{i+1})) \bigr) \bigr\lvert^2.
\end{eqnarray}
Note that for the explicit Euler scheme, as in eqn \eqref{eqn:forwardEulerBPTODE}, up to time $t_N$ we can infer $f_\theta$ only up to time $t_{N-1}$. Hence, there is a time lag in the learning which is not observed for the implicit Euler scheme.


The loss functionals in eqns \eqref{eqn:forwardEulerBPTODE} and \eqref{eqn:backwardEulerBPTODE} are trivially differentiable and acyclic (as the values of $t_i$ and $u(t_i)$ are just constants that have been taken from observations) as long as the graph representation of $f_\theta$ is differentiable and acyclic. Thus, if $f_\theta$ is represented by a differentiable and acyclic compute graph, the loss functionals $\hat{J}(\theta)$ can be optimised by SGD.

\subsection{Example using linear multistep integration approximation}

More sophisticated integration schemes than Backward or Forward Euler can be used to find a differentiable parametric representation of $\hat{J}(\theta)$. Linear multistep integral approximation schemes are briefly described here as they will be used for the numerical simulations presented in the next section of this paper. Any numerical scheme that is differentiable and representable by a directed acyclic compute graph when inserted into the loss functional could be used. Linear multistep methods are a convenient choice when the training data consists of observations of $u(t)$ that have been sampled at constant frequency.

From \cite{Hairer1993}, Adams-Moulton linear multistep integration of order $s=2$ can be used to approximate a trajectory of a dynamical system from time $a$ to time $b=a+2h$ for some $h \in \mathbb{R}$ as follows:
\begin{eqnarray}
    \hat{u}(b) = u(a+h) + h\left( \frac{5}{12}f_\theta(b,u(b)) + \frac{2}{3}f_\theta(a+h,u(a+h)) - \frac{1}{12}f_\theta(a,u(a)) \right).
    \label{eqn:adamsMoultonThree}
\end{eqnarray}

To derive the loss functional $\hat{J}(\theta)$, assume that training data observations of $u(t)$ are given by $\lbrace t_i, u(t_i) \rbrace_{i=1}^N$ and that the times $t_i$ are evenly spaced such that $t_i = (i-1)h$. Inserting eqn \eqref{eqn:adamsMoultonThree} into eqn \eqref{eqn:integralOptDiscreteFinal} gives the Adams-Moulton approximate loss functional ($m=3$, $p=2$):
\begin{eqnarray}
  \label{eqn:adamsMoultonLoss}
  \hat{J}_A(\theta) := \frac{1}{N-2} \sum_{i=1}^{N-2} \bigl\lvert u(t_{i+2}) - \hat{u}(t_{i+2})\bigr\lvert^2.
\end{eqnarray}

Note that the full Adams-Moulton integrator (defined in \cite{Hairer1993}) could also be used to derive a loss functional that approximates a trajectory discretisation using a series of interpolation points between the observations in the training set. For simplicity, only the method shown above (placing the evaluation points at the values in the training data set) is used in this paper.

\section{NUMERICAL ANALYSIS OF THE LORENZ--EMANUEL SYSTEM}
\label{sec:mainNumerical}

\subsection{Overview}

This section demonstrates the application of the parametric polynomial kernel regression technique to the model inference problem for a dynamical system using the discretisation detailed in the previous section of this paper. Simulations of the  Lorenz--Emanuel system (see \S 7.1 of \cite{clinton2010elegant}) were analysed. This dynamical system consists of $N$ variables, $u_i$ for $1 \leq i \leq N$, arranged periodically such that $u_{N+1} = u_1, u_0 = u_N$ and $u_{-1} = u_{N-1}$. Let the full set of variables be denoted by $u := \lbrace u_i\rbrace_{i=1}^N$. The Lorenz--Emanuel system can be highly chaotic, displaying sensitive dependence on initial conditions. The equations of motion for this system are:
\begin{eqnarray}
  \label{eqn:lorenzESys}
  \frac{du_i}{dt} = \left(u_{i+1} - u_{i-2} \right)u_{i-1} - u_i + F.
\end{eqnarray}
For the analysis in this section, the following parameters were adopted:
\begin{eqnarray}
  N = 8, \quad F = 5.
\end{eqnarray}
The parameter $F$ represents an external forcing term that prevents the energy in the system from decaying to zero. The value $F = 5$ was chosen to be high enough to cause sensitive dependence on initial conditions.

\subsection{Model inference training data and test description}

\begin{figure}[tb]
  \begin{center}
    \includegraphics{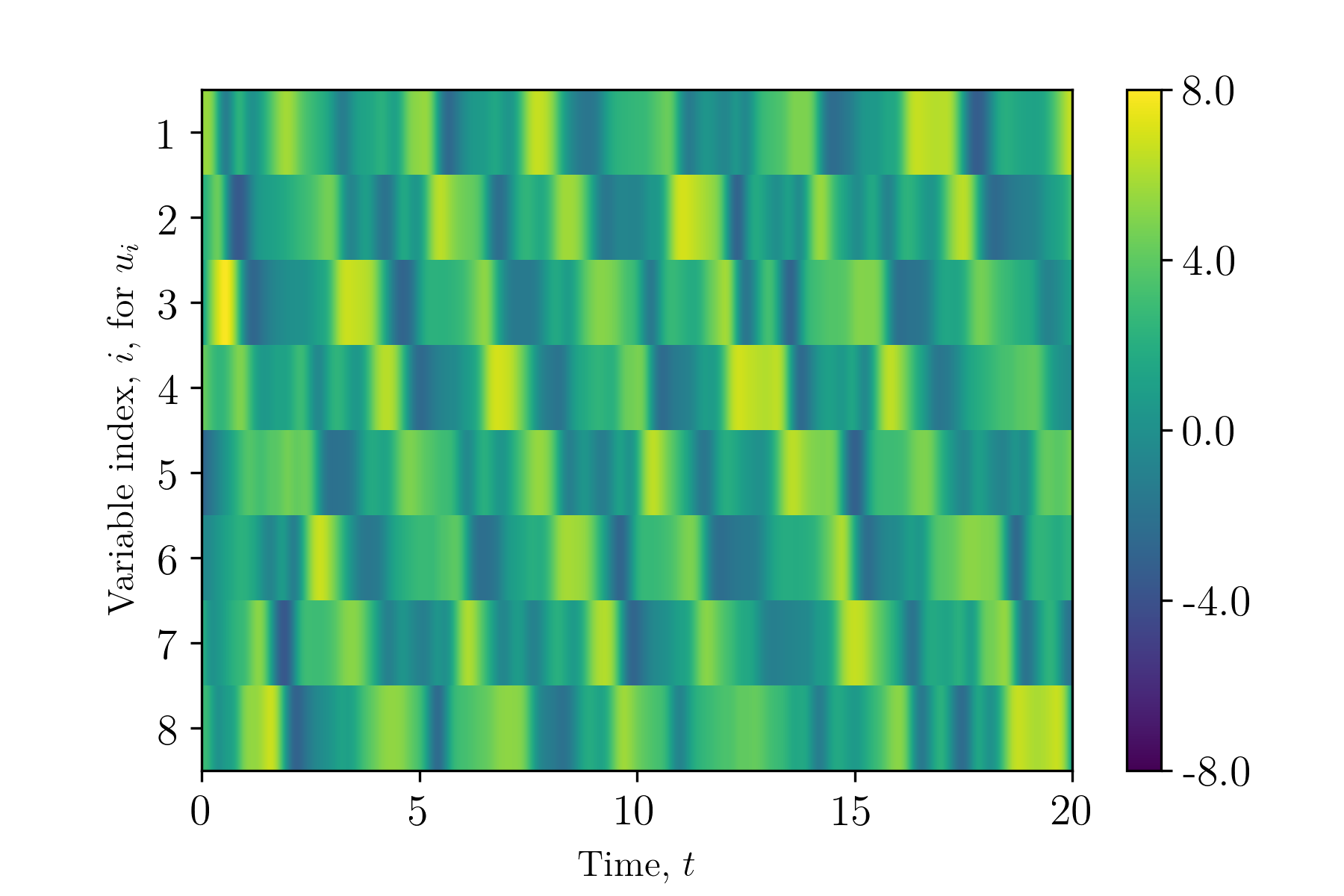}
    \caption{Lorenz--Emanuel system training data, generated using the model defined in eqn \eqref{eqn:lorenzESys}.}
    \label{fig:lorenzETrainingData}
  \end{center}
\end{figure}

Model inference was performed given the training data shown in fig \ref{fig:lorenzETrainingData}. The training data was generated using the SciPy solve\_ivp method \cite{scipy2001} with the `RK45' algorithm (variable 4th-5th order Runge-Kutta \cite{dormand1986}) and sampled at a rate of $1000$ samples per time unit for times $t=0$ to $t=20$. The initial values for the data were generated by sampling each $u_i$ independently from a normal distribution with mean $0$ and standard deviation $3$:
\begin{eqnarray}
  \label{eqn:initCondsEqn}
  u_i(t=0) \sim \mathcal{N}(\mu=0,\sigma=3).
\end{eqnarray}

The performance of the proposed method was tested by resampling new initial conditions from the same distribution in eqn \eqref{eqn:initCondsEqn} and comparing the outputs from the true simulation to simulations generated using an inferred model. All test simulations were again carried out using the SciPy solve\_ivp method with `RK45' integration \cite{scipy2001,dormand1986}.

We used the Adams-Moulton loss functional, $\hat{J}_A(\theta)$, in eqn \eqref{eqn:adamsMoultonLoss} to define the model inference task. The specific form of the inferred models is given in Section \ref{ssec:numericalExampleModel}. All models were implemented using Tensorflow \cite{tensorflow2015-whitepaper} and optimised with the Adam variant of SGD (see \cite{Kingma2014} for implementation details). A fixed optimisation training schedule was adopted in all cases and consisted of three phases, $P_1,P_2,P_3$. Each phase is described by an ordered pair $(I_i,\eta_i)$ where $I_i$ is the number of gradient descent iterations for that phase and $\eta_i$ is the `learning rate' parameter as in eqn \eqref{eqn:gradDescentExpand}. The training schedule adopted was:
\begin{eqnarray}
  \lbrace P_1 =(1000,0.1), P_2=(2000,0.01), P_3=(200,0.001) \rbrace.
\end{eqnarray}
It was found that this schedule was sufficient to minimise $\hat{J}_A(\theta)$ to approximately the maximum achievable precision for all models tested.

Note that the integrator used to generate trajectories (RK45) and that used for discretisation of the ODE trajectories (Adams-Moulton) are not the same. This was to demonstrate that any ODE solver can be used to generate simulations from the inferred model.


\subsection{Model representation with polynomial linearisations and kernels}
\label{ssec:numericalExampleModel}

To complete the specification of the problem, the basic form of $f_\theta$ must be provided. If the form of the dynamical system equations are known beforehand, this information can be used to simplify the analysis. If no information is available, a search over different types of compute graph architectures must be conducted (as in \cite{stanley2002}). For this demonstration, only a polynomial structure is assumed. This is a reasonable assumption that one could make when investigating general interdependent data observations from a dynamical system without any other prior knowledge, as a number of systems have such a structure \cite{clinton2010elegant}.

For this inference task, the exact form of the polynomial couplings between the various $u_i$ were not provided to the compute graph. Instead, two types of polynomial nonlinearities were tested. First, a linear combination of all second-order polynomial terms that could be constructed using each of the $u_i$ terms was considered, that is, equations of the form
\begin{eqnarray}
  \label{eqn:polyFeature}
  \frac{d \hat{u}_i}{dt} = f^i_\theta(u) = \sum_{k=1}^N \sum_{j=1}^{k} \alpha^i_{kj} u_k u_j + \beta_k^i u_k + \gamma_i
\end{eqnarray}
for each $i \in [1,\dots,N]$.
The parameters are $\gamma_i \in \mathbb{R}$, $\beta_k^i \in \mathbb{R}^N$, $\alpha^i_{kj} \in \mathbb{R}$ for $i \in [1,\dots,N]$, $k = \left[ 1, \cdots, N \right]$, $j = \left[ 1, \cdots, k \right]$. This sort of polynomial is of the traditional form used for polynomial chaos expansions (see \cite{sudretHDR}).

Second, the parametric polynomial kernel method introduced in this paper and defined in eqn \eqref{eqn:secondOrderPolyModel} with dimensions $D=E=8$ was used to represent $f_\theta$. Values of $M=60,80$ and $100$ were tried to test the effect of this parameter on the accuracy of the results.



\subsection{Results}

Stochastic Gradient Descent, combined with ODE trajectory discretisation, was successfully applied to model inference for the Lorenz--Emanuel system in eqn \eqref{eqn:lorenzESys}. Our parametric kernel model gave the best accuracy on the inference task. Importantly, the kernel model was able to be tuned to higher accuracies by increasing the number of weights used, $M$. Although increasing $M$ increases the number of total parameters to be optimised, this trade off may be worthwhile depending on the particular problem.

\begin{figure}[tb]
  \begin{center}
    \includegraphics{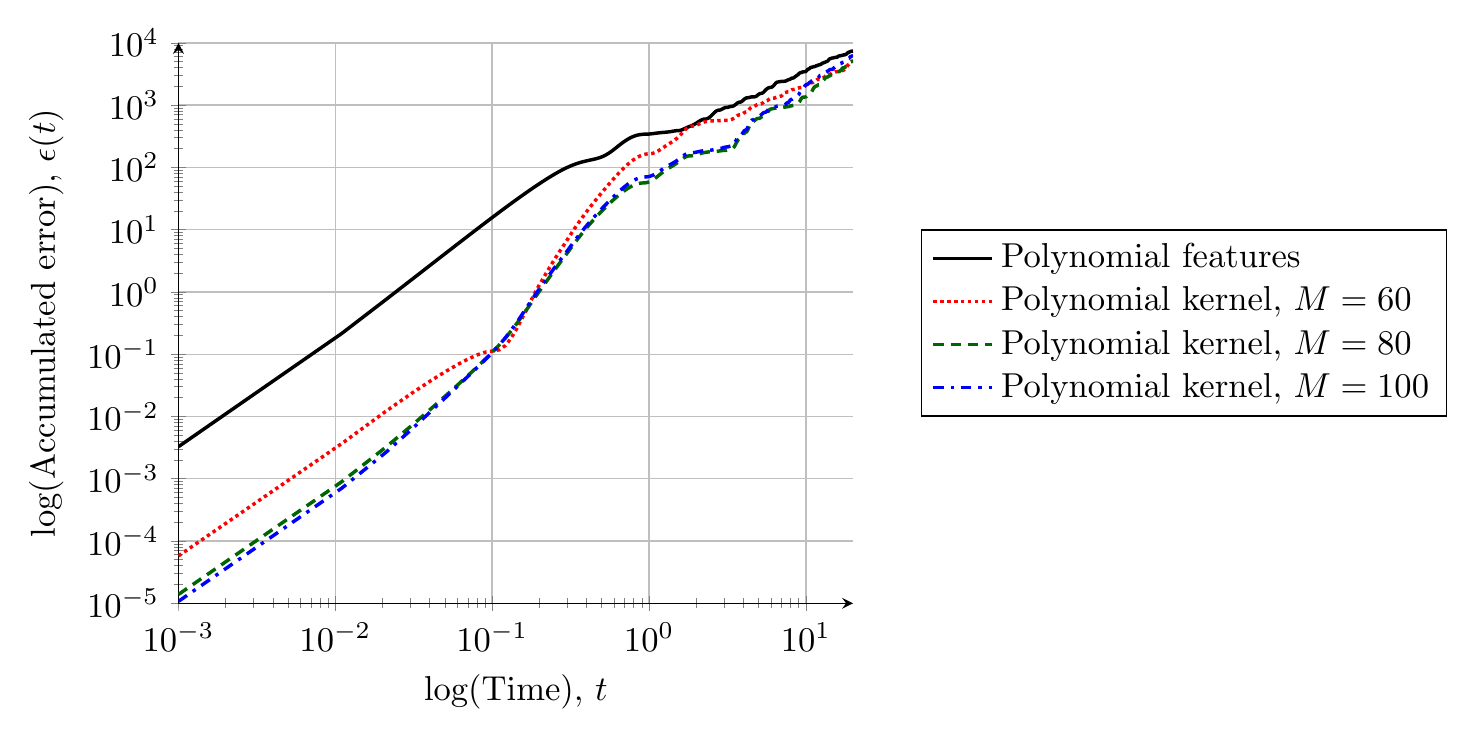}
    \caption{Lorenz--Emanuel system error vs time. Errors are calculated as per eqn \eqref{eqn:lorenzError}.}
    \label{fig:lorenzEErrVsTime}
  \end{center}
\end{figure}

The performance of the different models is shown in fig \ref{fig:lorenzEErrVsTime}. The accumulated error, $\epsilon(t)$, was calculated as the sum of squared errors from the true model:
\begin{eqnarray}
  \epsilon(t=0) &=& 0, \\
  \label{eqn:lorenzError}
  \epsilon(t+h) &=& \sqrt{\left((u(t+h)-\hat{u}(t+h)\right)^2} + \epsilon(t),
\end{eqnarray}
where $h = 0.001$ (matching the training data sampling rate of $1000$ samples per time unit). The errors were calculated for the polynomial feature model in eqn \eqref{eqn:polyFeature} and the polynomial kernel model in eqn \eqref{eqn:secondOrderPolyModel} for $M=60$, $M=80$ and $M=100$.

From fig \ref{fig:lorenzEErrVsTime}, the direct polynomial feature mapping had the worst accuracy. The parametric kernel method was able to track the system evolution more accurately. In all cases, the inferred models were able to maintain a small inference error at times up to at least an order of magnitude greater than the training data sampling rate.

\begin{figure}[tb]

  \begin{subfigure}[b]{\textwidth}
    \begin{center}
      \includegraphics{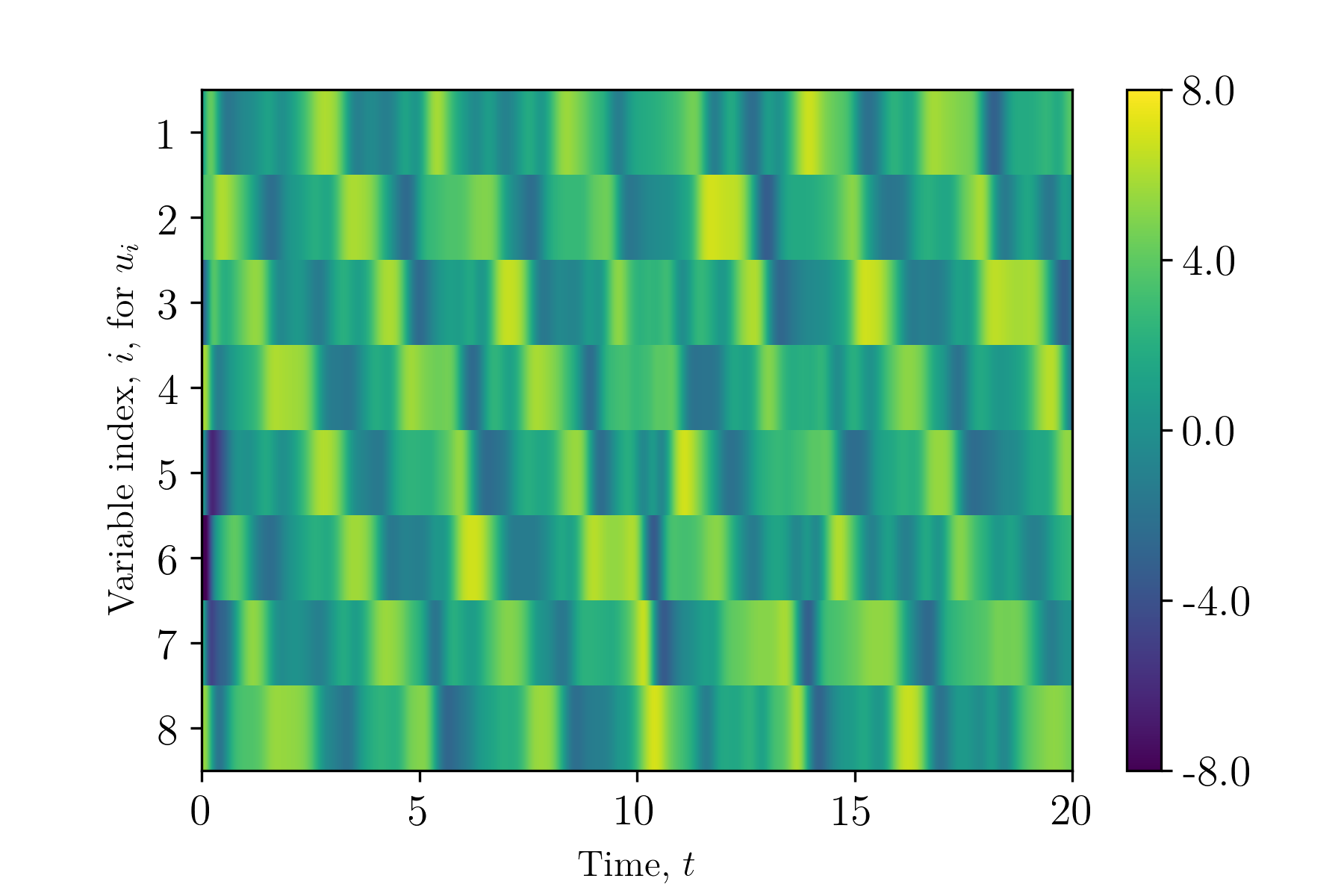}
      \caption{Data trace from true model, $u(t)$.}
    \end{center}
  \end{subfigure}
  \begin{subfigure}[b]{\textwidth}
    \begin{center}
      \includegraphics{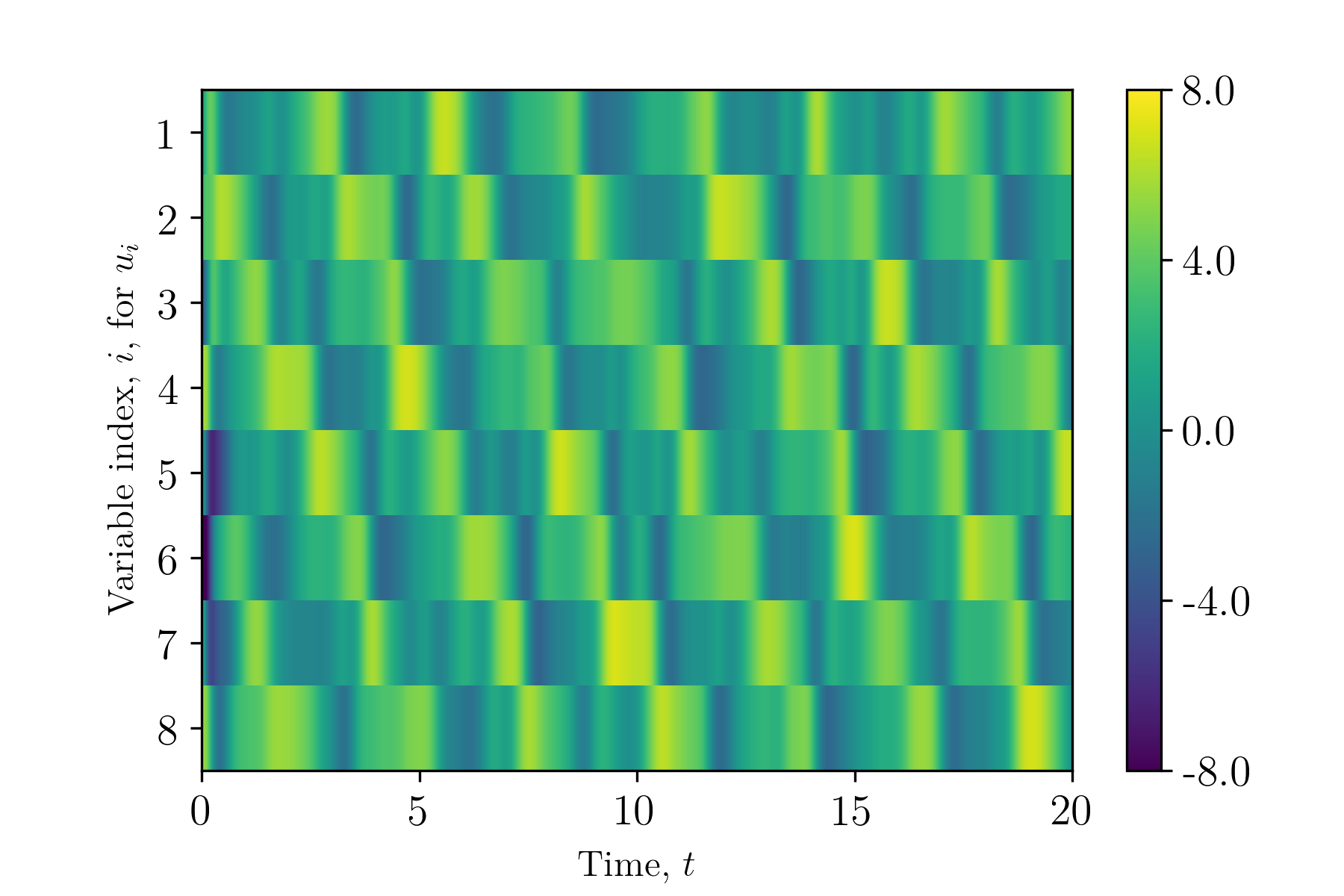}
      \caption{Data trace from inferred parametric polynomial kernel model, $\hat{u}(t)$, with $M=100$.}
    \end{center}
  \end{subfigure}
  \begin{center}
    \caption{Comparison of output traces for the Lorenz--Emanuel system, defined in eqn \eqref{eqn:lorenzESys}: (a) true system simulation, $u(t)$, and (b) most accurate inferred model, $\hat{u}(t)$. The inferred model structure is given by the parametric polynomial kernel in eqn \eqref{eqn:secondOrderPolyModel} for $M=100$.}
    \label{fig:lorenzEExampleTrace}
  \end{center}
\end{figure}

Performance on the model inference task for the polynomial kernel method defined by eqn \eqref{eqn:secondOrderPolyModel} with $M=100$ is demonstrated in fig \ref{fig:lorenzEExampleTrace}. This pair of figures shows a comparison between the true model output, $u(t)$, and inferred model output, $\hat{u}(t)$. From fig \ref{fig:lorenzEExampleTrace}, it can be seen that the overall structure of the equations is captured by the inferred model. Due to the chaotic nature of the system being analysed, once a few errors accumulate, the true and inferred models diverge rapidly.

\subsection{Discussion}

The parametric polynomial kernel method was able to infer the hidden ODE model with good accuracy given a fixed set of training data. The accumulated errors grow quickly with time. This is reasonable considering the chaotic nature of the Lorenz--Emanuel system. A more mathematically rigorous stability analysis of the numerical scheme would be interesting but is beyond the scope of this paper. A number of possible variations on the numerical example presented could be analysed in future work. For instance, the type of integration method used, the sampling rate of the data, and the effect of different amounts of training data would all be interesting to investigate.

\section{CONCLUSIONS}


This paper presented a parametric form of polynomial kernel regression, as well as numerical case studies. In particular, the proposed method was applied to the model inference problem for a chaotic dynamical system. Our parametric polynomial kernel method was able to harness the power of kernelised regression without the cubic computational complexity typically incurred by nonparametric polynomial regression, thereby avoiding the curse of dimensionality. Although the method was successfully applied to a test problem, more work will be required to fully understand how best to apply parametric polynomial kernels to real world (rather than simulated) data. As is the case in all regression models, some form of regularisation would need to be included to address overfitting and observational noise.

It was assumed for the analysis in this paper that it was known a priori that only certain polynomial couplings are present. Using the wrong polynomial order in the model expansion was found to cause convergence difficulties. This is also the case in nonparametric kernel regression (see \cite{murphy2012machine} and the example in fig \ref{fig:fittedPolynomialTest}). As such, this is not considered a serious limitation of the method in that it is possible to test a few different sets of model forms when attempting to find a good fit to a data set. Bayesian model selection methods could be applied to formally assess the quality of different polynomial kernel model dimensions.

It is worth noting that direct projection onto polynomial features was found to perform poorly compared to the polynomial kernel method. Although stochasticity was not considered in this paper, it is quite possible that this finding will impact standard techniques frequently employed for Uncertainty Quantification. A kernel representation of the type introduced in this paper applied to Gaussian and other stochastic features may be useful for improving standard polynomial chaos methods (which are described in \cite{sudretHDR}).


The search for effective compute graph architectures remains a problem that plagues all methods attempting to learn hidden function structures without inserting large amounts of prior knowledge into the inverse problem. Scaling to very high-dimensional problems would be an interesting challenge. Given the partial decoupling from the curse of dimensionality that gradient descent methods can provide, it is hoped that the techniques presented in this paper would be suitable for model inference on large scale dynamical systems in the future.


\section{ACKNOWLEDGEMENTS}

This project has received funding from the European Research Council (ERC) under the European Union's Horizon 2020 research and innovation programme, grant agreement No 757254 (SINGULARITY) and a Lloyds Register Foundation grant for the Data-Centric Engineering programme at the Alan Turing Institute.


\end{document}